\documentclass{article} 
\usepackage{iclr2025_conference,times}

\usepackage{amsmath,amsfonts,bm}

\def\eqref#1{equation~\ref{#1}}

\def\1{\bm{1}}

\DeclareMathAlphabet{\mathsfit}{\encodingdefault}{\sfdefault}{m}{sl}
\SetMathAlphabet{\mathsfit}{bold}{\encodingdefault}{\sfdefault}{bx}{n}

\usepackage{hyperref}
\usepackage{url}
\usepackage{graphicx}
\usepackage{subfigure}
\usepackage{caption}
\usepackage{enumerate}

\title{Accelerating Transformer Inference and Training with 2:4 Activation Sparsity}

\iclrfinalcopy

\author{
Daniel Haziza \And Timothy Chou \And Dhruv Choudhary \And Luca Wehrstedt \And Francisco Massa \And Jiecao Yu \And Geonhwa Jeong \And Supriya Rao \And Patrick Labatut  \And Jesse Cai \\
Meta \\
1 Hacker Way, Menlo Park, CA 94025 \\
\texttt{\{dhaziza,timchou,choudharydhruv,lcw\}@meta.com} \\
\texttt{\{fmasssa,jiecaoyu,geonhwa,supriyar,palabatut,jessecai\}@meta.com}\\
}

\begin{document}

\maketitle

\begin{abstract}

In this paper, we demonstrate how to leverage 2:4 sparsity, a popular hardware-accelerated GPU sparsity pattern, to activations to accelerate large language model training and inference. Crucially we exploit the intrinsic sparsity found in Squared-ReLU activations to provide this acceleration with \textbf{no accuracy loss}. Our approach achieves up to 1.3x faster Feed Forward Network (FFNs) in both the forwards and backwards pass. This work highlights the potential for sparsity to play a key role in accelerating large language model training and inference. 
\end{abstract}

\section{Introduction \& Related Work}

The rapid growth of Large Language Models (LLMs) in recent years has been driven by a corresponding surge in GPU FLOPs. While some of this gain has come from hardware improvements, a large portion has come from reducing operator precision from $32 \rightarrow 16 \rightarrow 8 $ bits. However, as we approach the $0$ bit quantization asymptote, researchers have turned to alternative methods to reduce LLM compute. One promising approach is sparsity, which seeks to avoid unnecessary computation to accelerate the model. 

Existing work on sparsity for LLMs has primarily focused on inference.  SparseGPT \citep{frantar2023sparsegpt} and others have shown that existing dense LLM weights can be made sparse, either through post training pruning \citep{sun2023wanda, zhang2024plugandplay} or fine-tuning \citep{kurtic2023sparse}, enabling subsequent inference acceleration. Sparsity has also been applied to activations \citep{liu2024trainingfreeactivationsparsitylarge, lee2024cats, zhang2024relu2winsdiscoveringefficient} to speed up batch size=1 decode inference workloads via selective weight loading. However, these approaches are unable to provide speedups during LLM training or prefill, where we experience compute-bound workloads. 

\cite{mozaffari2024slope}, \cite{hu2024sste} were able to accelerate LLM training with 2:4 sparsity, but applied to the model weights. Moreover, in all of these works, the final sparse model comes with a small reduction in model accuracy compared to the dense model.

In this paper, we investigate the application of 2:4 sparsity to accelerate both LLM training and inference by sparsifying model activations. Crucially, we exploit the intrinsic sparsity found in certain activation functions, which is not present in the dense model weights. Our main contributions are as follows:

\begin{enumerate}[(1)]
\item A recipe to accelerate Squared-RELU LLM training and inference with 2:4 sparsity.
\item Accuracy experiments showing minimal performance degradation when using our emulated recipe compared to the dense baseline.
\item Kernel benchmarks showing up to 1.3x faster FFNs when using 2:4 sparsity. 
\end{enumerate}

\textbf{Squared ReLU: } Recently, SwiGLU = $(Swish_{\beta}(XW_1) \odot XW_3)W_2$ \citep{shazeer2020glu} has been used as the activation function in the Feed-Forward Network blocks (FFN) of multiple LLMs \citep{touvron2023llama} \citep{liu2024deepseek}. However, \citep{so2021primer} suggests that Squared-ReLU  = $({\max(0, XW_1)}^2)W_2$ is at least as good as SwiGLU, which we confirmed in our experiments (Table\ref{accuraccy-results}). Larger models have since been trained successfully with Squared-ReLU \citep{adler2024nemotron}.

\begin{figure}[h]
\begin{center}
\includegraphics[width=0.75\linewidth]{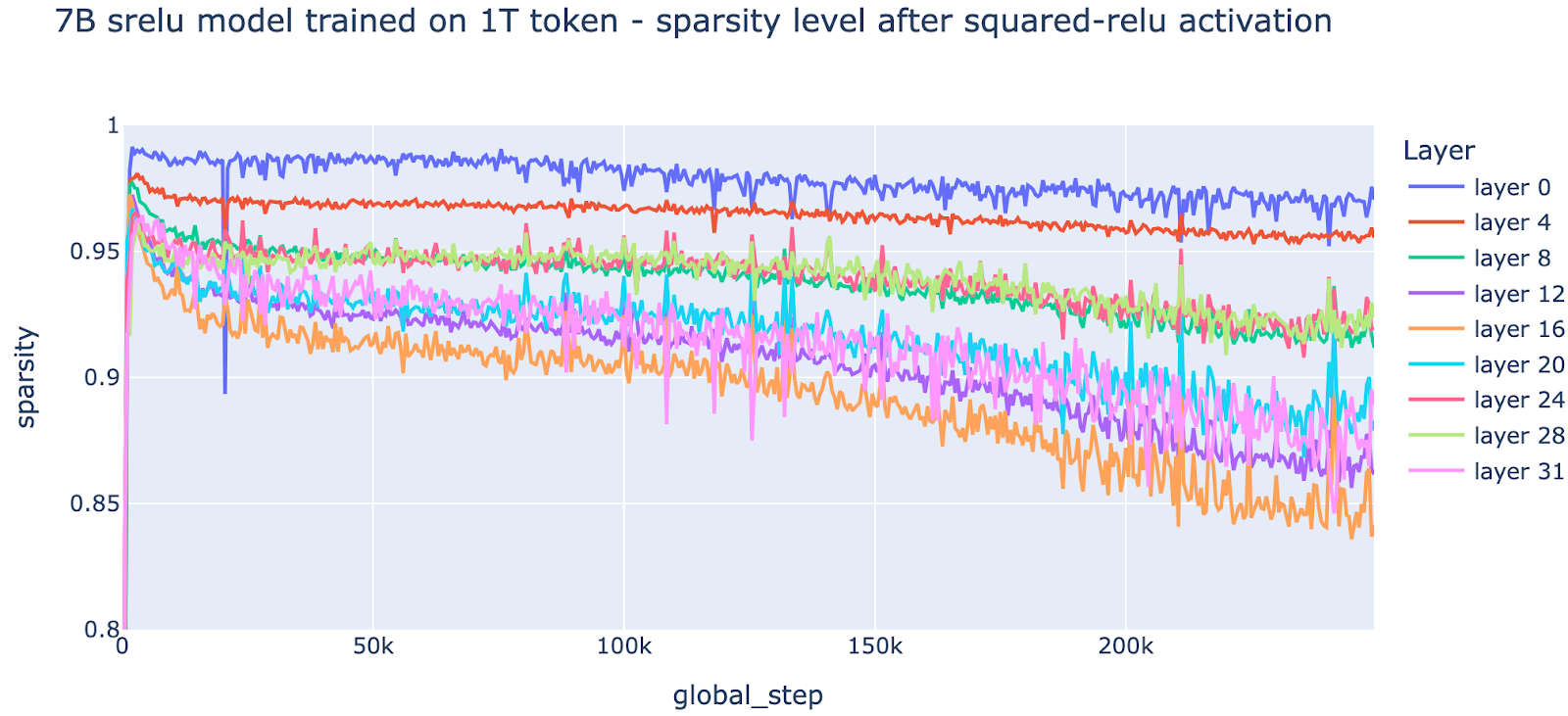}
\end{center}
\captionsetup{font=small, textfont=it, width=0.9\linewidth}
\caption{\textit{We plot the sparsity level progression of different FFN layers over a training run. Replacing SwiGLU with Squared-ReLU does not impact the model accuracy (See Table\ref{accuraccy-results}), and makes the FFN activations highly sparse during training \& inference (84-98\% sparse for this model).}}
\label{fig:sparsity_level}
\end{figure}

While the model accuracy remains similar, switching the activation function to Squared-ReLU has an interesting side-effect: activation sparsity naturally emerges during training. Intuitively, this makes sense as $Swish_{\beta}(x) = \frac{x}{1 + e^{-\beta x}}$ will be 0 if and only if $x=0$ whereas $ReLU^2 = {\max(0, x)}^2$ maps any $x < 0 $ to 0. In theory, with normal centered inputs, we expect to see 50\% sparsity in the activations - which is what we observe at model initialization. In practice, however, we observe that the sparsity level rapidly increases during training, reaching 85-98\% depending on the layer (Figure\ref{fig:sparsity_level}). While we do not offer an explanation for the emergence of this phenomenon, this means that a Squared ReLU model spends a large amount of GPU time and energy multiplying zeros. \footnote{While we only examine Squared-RELU models in this paper, this approach can be adapted to any activation function that achieves similar sparsity levels.}

\begin{figure}[h]
\begin{center}
\includegraphics[width=\linewidth]{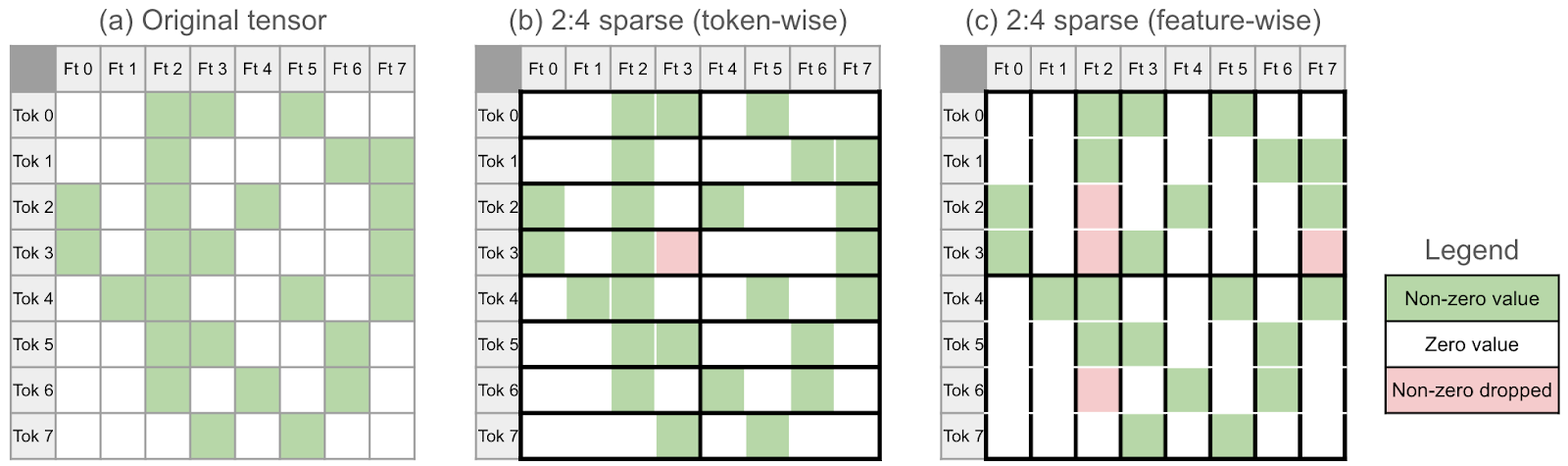}
\end{center}
\captionsetup{font=small, textfont=it, width=0.9\linewidth}
\caption{\textit{Given an activation tensor A of shape $[seqlen, features]$ (Figure 2a), we can accelerate the computation of $AB$ if $A$ is 2:4-sparse token-wise (Figure 2b), or $A^TB$ if $A$ is 2:4-sparse feature-wise (Figure 2c). To respect the 2:4 sparsity constraint, it is possible that some values need to be dropped when the tensor A is sparsified (in red). The more sparse the original tensor is, the less likely we are to drop values - if no values are dropped, the calculation is exact. }}
\label{fig:spgemm}
\end{figure}

\textbf{Hardware-accelerated (2:4) sparsity: } To accelerate this model, we leverage sparsity. Modern GPUs have specialized units for matrix multiplication (TensorCores), and these units can only accelerate a single sparsity format: 2:4 sparsity \citep{mishra2021accelerating}. A tensor is 2:4 sparse if for every 4 consecutive values, at most 2 are non-zeros (Figure\ref{fig:spgemm}). If a tensor $A$ is 2:4-sparse (50\% sparsity), it is theoretically possible to compute $AB$ 2x faster, because we skip every other multiplication - in practice we see 1.5x-1.7x speedups for 2:4 FP8 matrix multiplication on H100 compared to dense FP8 depending on the matrix shape.

\section{Approach}
\label{Approach}

Inspired by \citet{jeong2024abstracting}, we aim to approximate the second \footnote{Note that the first linear layer cannot be accelerated as its inputs are not sparse. Further savings could be achieved if the residual can be made sparse.} matrix multiply of the FFN, $Y_2W_2$, with a 2:4 sparse matrix multiply, as $Y_2 = ReLU^2(X_1W_1)$ and is mostly zero. If more than 2 elements are non-zeros in a block of 4 in $Y_2$, we naively keep the 2 highest magnitude scalars and drop the others. In practice, $\sim$1\% of non-zero values are dropped, but this can vary depending on the sparsity level.

\begin{figure}[!htb]
\begin{center}
\includegraphics[width=0.5\linewidth]{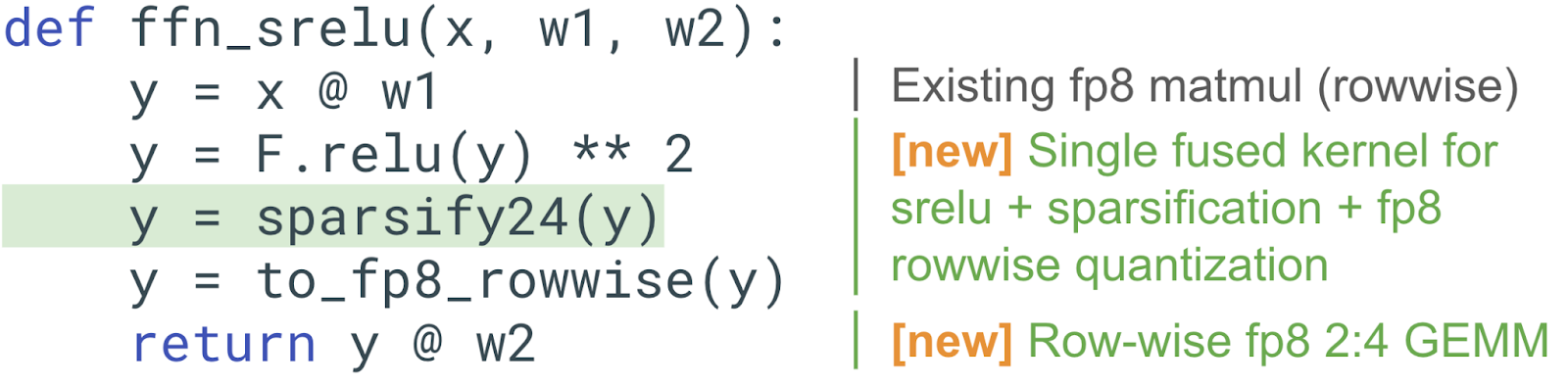}
\end{center}
\captionsetup{font=small, textfont=it, width=0.8\linewidth}
\caption{\textit{Pseudo-code for our proposed replacement FP8 Squared-ReLU FFN.}}
\end{figure}

We implemented 2 new kernels, so that the entire FFN forward pass can be implemented with only 3 kernels: the existing FP8 matmul for the first linear layer, a new kernel that applies the activation, the sparsification, and FP8 quantization, and another new 2:4 GEMM kernel that supports FP8 rowwise scaling. To sparsify the activations we adapt the fast sparsification routines outlined in \cite{pytorchAcceleratingNeural} \footnote{These routines have been open sourced in \href{https://github.com/pytorch/ao/tree/main/torchao/sparsity/training}{torchao} \citep{torchao}}

\begin{figure}[!htb]
\begin{center}
\subfigure{\includegraphics[width=.45\linewidth]{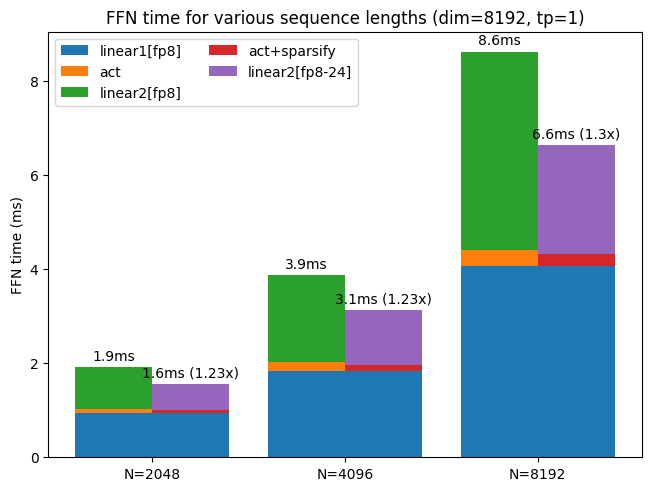}}
\subfigure{\includegraphics[width=.45\linewidth]{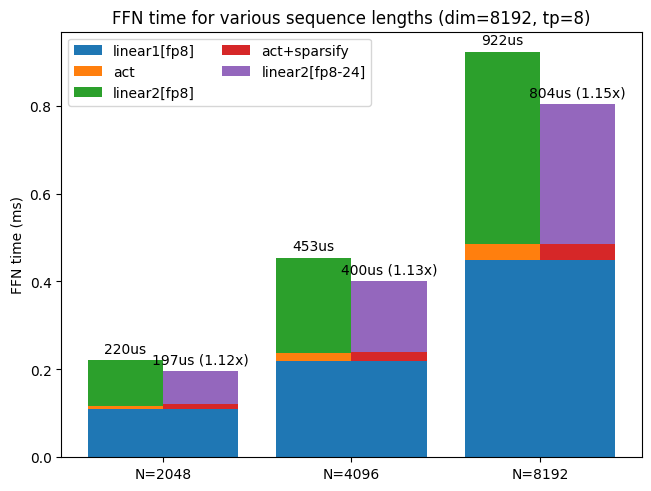}}
\end{center}
\captionsetup{font=small, textfont=it, width=0.9\linewidth}
\caption{\textit{Our kernels can accelerate the FFN forward pass up to 30\%, depending on the model dimension and batch size. Larger models or larger batch sizes lead to higher speedups. All FP8 matrix multiplications are done with row-wise scaling, to match the baseline training recipe.}}
\label{fig:forwards_benchmark}
\end{figure}

Notably, this also provides inference speedups in the compute bound regime (Figure\ref{fig:forwards_benchmark}), such as during prefilling, with speculative decoding, or higher batch sizes. This is different from existing activation sparsity work \citep{liu2024trainingfreeactivationsparsitylarge, lee2024cats} that aims at reducing the amount of weights loaded from memory, which can only accelerate batch size=1 decode.

\begin{figure}[h]
\begin{center}
\includegraphics[width=0.9\linewidth]{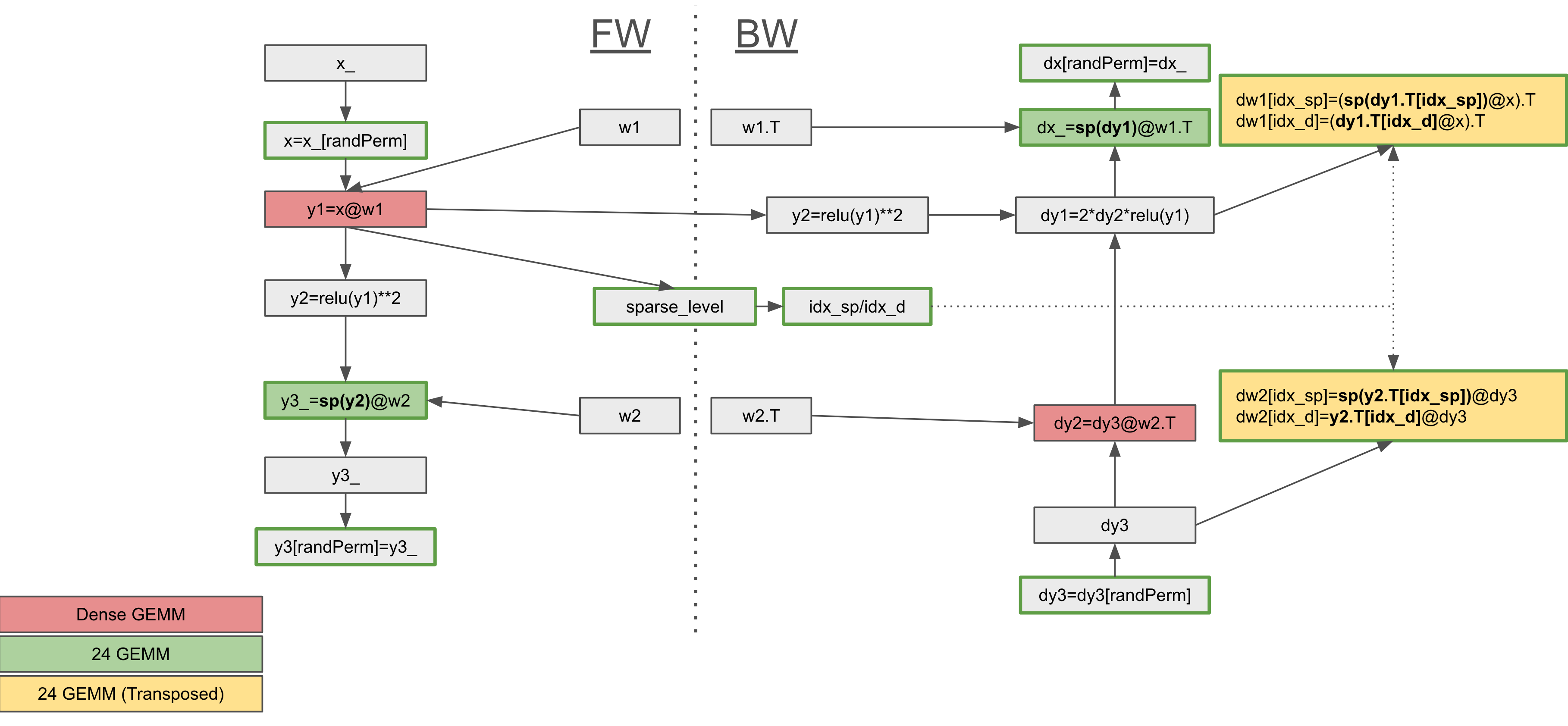}
\end{center}
\captionsetup{font=small, textfont=it, width=0.9\linewidth}
\caption{\textit{The compute graph for a Squared-ReLU FFN during training. As $ReLU^2(y_1)$ is highly sparse, any matrix multiplication involving $y_2$ or $dy_1$ can be accelerated. In total, 4  matrix multiplications out of 6 can be accelerated, when considering both the forward and backward passes.}}
\end{figure}

Unfortunately the sparse GEMMs in the backwards pass are more difficult to accelerate. NVIDIA GPUs can only accelerate sparsity along the reduction dimension. During the backwards pass when we calculate $\frac{\partial W_2}{\partial y_3} = y_2^T\frac{\partial y_3}{\partial L}$ we reduce on the feature dimension, and therefore $y_2$ must be sparsified feature-wise (Figure\ref{fig:spgemm}b) to leverage hardware acceleration. Naively sparsifying as we did in the forwards pass yields poor accuracy, due to the following reasons: 

\textbf{Some individual features are not  sufficiently sparse (e.g. 20\% sparsity only) and are unable to be sparsified.} Therefore, we introduce \textit{Optimization 1) split the sparse tensors into 2 tensors: a tensor with 95\% of the features, that can be 2:4 sparsified feature-wise , and a tensor with the remaining of the values, containing the non-sparse features}. This means that we execute 2 GEMMs: a 2:4 sparse GEMM with 95\% of the flops, and a dense one with 5\% of the flops. The dense GEMM in the forward pass, $Y_1=XW_1$ can compute the column level sparsity in the epilogue for free, while
a cheap argsort kernel is then used to partition features into sparse and non-sparse features.

\textbf{Features are highly correlated among consecutive tokens.} For instance, a feature might be 99\% sparse, but non-zero on 10 consecutive tokens. For this reason, we apply \textit{Optimization 2) a fixed permutation to the tokens before entering the FFN and shuffle them back after}. These tokens permutations can be fused into the existing add/quantize/normalize operations before and after the FFN. In fact, it is possible to implement all the pointwise operations of the backward pass in a single fused kernel, including the transposition/FP8 scaling, and the split of tensors in dense \& sparse counterparts.

Finally, we sparsify feature-wise on top of the token-wise sparsification mask computed in the forward pass to maintain training stability, as sparsifying feature-wise would otherwise allow for some values that are dropped in the forwards pass to reappear. We also train the models densely for the first 1k iterations (warmup), as the sparsity at initialization is roughly 50\%, and few training steps are required to increase to 90\%+ sparsity.

\section{Experimental Results}
\label{Experiments}

We are still actively developing the kernels and fusions outlined above to show end-to-end speedups for training but have 1) \textbf{accuracy experiments on LLM pretraining that show no accuracy loss} and 2) \textbf{kernel benchmarks of the FFN forwards pass and backwards split-GEMM that show up to a 1.3x speedup}. We advocate for further scaling of sparse transformers to validate the approach with larger models, as well as exploration into why Squared-ReLU models have such high levels of intrinsic sparsity.

\begin{table}[!htb]

\begin{center}
\begin{tabular}{ll}
\hline 
\multicolumn{1}{c}{\bf Experiment}  &\multicolumn{1}{c}{\bf
Final Perplexity}\\
\hline  
Dense training (SwiGLU)               & 2.654 \\
Dense training (Squared-ReLU)                & 2.651 \\
2:4 recipe (5\% of the features dense in BW) & 2.652 \\
2:4 - no warmup                        & 2.657 \\
2:4 - naively sparsify backward GEMMs     & 2.682 \\
2:4 - no permuting rows                & 2.919 (plateaus very early) \\
2:4 - no sparsify $y_1$ in BW pass       & 3.735 (diverges after step $\sim$42k) \\
\hline
\end{tabular}
\captionsetup{font=small, textfont=it, width=0.9\linewidth}
\caption{LLM pretraining accuracy. These experiments were ran on an 1.5B LLM trained for 63B tokens on DCLM with emulated 2:4 sparsity. Please see Appendix for more details. }
\label{accuraccy-results}

\end{center}
\end{table}

\begin{figure}[!htb]
\begin{center}
\includegraphics[width=0.5\linewidth]{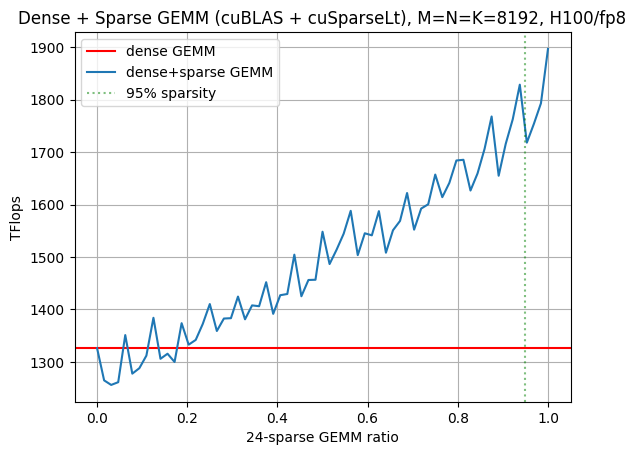}
\end{center}
\captionsetup{font=small, textfont=it, width=0.9\linewidth}
\caption{\textit{split GEMM TFLOPs across different split ratios. We decompose the feature-wise sparse GEMMs found in the backwards pass as 95\% 2:4-sparse and 5\% dense, which incurs a small performance hit compared to having a fully 2:4 sparse GEMM, but is still faster than a fully dense GEMM. Operand A contains 90\% of zeros.}}
\label{fig:training_benchmark}
\end{figure}

\newpage

\bibliographystyle{iclr2025_conference}
\bibliography{iclr2025_conference}

\appendix
\section{Appendix}

\begin{table}[!htb]
\label{hparam}

\captionsetup{font=small, textfont=it, width=0.9\linewidth}
\caption{LLM training hyperparameters. We use the same hyperparameters for training the SwiGLU and Squared-ReLU variants, and modify the FFN hidden dimension to ensure we have the same total number of model parameters}
\begin{center}
\begin{tabular}{lll}
\hline 
\multicolumn{1}{c}{\bf Hyperparameter}  &\multicolumn{1}{c}{\bf
Value for 1B model} & \multicolumn{1}{c}{\bf
Value for 7B model}\\
\hline  
Training dataset               & DCLM  & DCLM\\
Number of GPUs & 64xH100 & 256xH100 \\
Global batch size (tokens) & 1M & 2M \\
Training iterations &  35k (37B tokens) & 100k (209B tokens) \\
Optimizer &  AdamW ($\beta_1=0.9$, $\beta_2=0.95$) & AdamW ($\beta_1=0.9$, $\beta_2=0.95$) \\
Learning rate & $3e-3$ & $1e-3$ \\
Learning rate schedule & 5,000 linear warmup steps + cosine & 2,000 linear warmup steps + cosine\\
Gradient clipping & $1.0$ & $1.0$ \\
Weight decay & $0.033$ & $0.1$ \\
Model dimension & $2048$ & $4096$ \\
FFN hidden dimension & $8192$ & $16384$ \\
Number of layers & $25$ & $32$ \\
Number of attention heads & $16$ & $32$ \\
\hline
\end{tabular}
\end{center}
\end{table}

\end{document}